\documentclass{article}

\usepackage{pdflscape} 
\usepackage{graphicx}
\usepackage[sglblindworkshop, final]{neurips_2025}
\workshoptitle{Machine Learning for the Physical Sciences}

\usepackage[utf8]{inputenc} 
\usepackage[T1]{fontenc}    
\usepackage{hyperref}       
\usepackage{url}            
\usepackage{booktabs}       
\usepackage{amsfonts}       
\usepackage{nicefrac}       
\usepackage{microtype}      
\usepackage{xcolor}         

\title{Connecting the Dots: A Machine Learning Ready Dataset for Ionospheric Forecasting Models}

\author{%
Linnea M.~Wolniewicz \\
  Department of Information and Computer Science \\
  University of Hawai`i at Mānoa, USA \\
  \texttt{linneamw@hawaii.edu} \\
  \And
  Halil S.~Kelebek \\
  Department of Engineering Science \\
  University of Oxford, UK \\
  \texttt{halil@robots.ox.ac.uk} \\
  \And
  Simone Mestici \\
  Università degli Studi di Roma Sapienza \\
  Rome, Italy \\
  \texttt{simone.mestici@uniroma1.it} \\
  \And
  Michael D.~Vergalla \\
  Free Flight Research Lab \\
  Sunnyvale, USA \\
  \texttt{mike@freeflightlab.org} \\
  \And
  Giacomo Acciarini \\
  European Space Agency (ESA) \\
  \texttt{giacomo.acciarini@esa.int} \\
  \And
  Bala Poduval \\
  University of New Hampshire \\
  \And
  Olga Verkhoglyadova \\
  NASA Jet Propulsion Laboratory \\
  \And
  Madhulika Guhathakurta \\
  NASA Headquarters \\
  \texttt{madhulika.guhathakurta@nasa.gov}
  \And
  Thomas E. Berger \\
  Space Weather Technology, Research, and Education Center  \\
  University of Colorado Boulder \\
  \texttt{Thomas.Berger@colorado.edu} \\
  \And
  Atılım Güneş Baydin \\
  Department of Computer Science \\
  University of Oxford, UK \\
  \texttt{gunes@robots.ox.ac.uk} \\
  \And
  Frank Soboczenski \\
  Department of Computer Science \\
  University of York \& King’s College London, UK \\
  \texttt{frank.soboczenski@york.ac.uk} \\
}

\begin{document}

\maketitle

\begin{abstract}

Operational forecasting of the ionosphere remains a critical space weather challenge due to sparse observations, complex coupling across geospatial layers, and a growing need for timely, accurate predictions that support Global Navigation Satellite System (GNSS), communications, aviation safety, as well as satellite operations. 
As part of the 2025 NASA Heliolab, we present a curated, open-access dataset that integrates diverse ionospheric and heliospheric measurements into a coherent, machine learning-ready structure, designed specifically to support next-generation forecasting models and address gaps in current operational frameworks.
Our workflow integrates a large selection of data sources comprising Solar Dynamic Observatory data, solar irradiance indices (F10.7), solar wind parameters (velocity and interplanetary magnetic field), geomagnetic activity indices (Kp, AE, SYM-H), and NASA JPL's Global Ionospheric Maps of Total Electron Content (GIM-TEC). We also implement geospatially sparse data such as the TEC derived from the World-Wide GNSS Receiver Network and crowdsourced Android smartphone measurements. 
This novel heterogeneous dataset is temporally and spatially aligned into a single, modular data structure that supports both physical and data-driven modeling. Leveraging this dataset, we train and benchmark several spatiotemporal machine learning architectures for forecasting vertical TEC under both quiet and geomagnetically active conditions.
This work presents an extensive dataset and modeling pipeline that enables exploration of not only ionospheric dynamics but also broader Sun-Earth interactions, supporting both scientific inquiry and operational forecasting efforts.

\end{abstract}

\section{Introduction}

Modern society is reliant on complex technological infrastructures, such as space-based navigation and communications systems, Low Earth Orbit (LEO) satellite constellations, aviation networks, and power grids, all of which are highly susceptible to disruptions caused by solar activity. Solar flares, coronal mass ejections, and energetic particles not only represent relevant risks to space operations but can also trigger geo-effective disturbances that directly impact life on Earth \cite{Berger2020}. The complex coupling between solar activity, the Earth's magnetosphere and the ionosphere-thermosphere systems drives geomagnetic storm events, capable of disrupting satellite operations, degrading Global Navigation Satellite Systems (GNSS) accuracy, compromising radio communicatons, and even precipitating power grid blackouts \cite{Kintner1976,Kataoka2022,Pulkkinen2017}. 

For these reasons, the past decades have seen a marked increase in missions monitoring near-Earth space. For instance, NASA's Tandem Reconnection and Cusp Electrodynamics Reconnaissance Satellites (TRACERS) mission \cite{Petrinec2025}, launched in 2025, deploys two satellites to study solar wind interactions in the polar cusp, aiming to enhance forecasting of geomagnetic storms. ESA's forthcoming Vigil mission \cite{Eastwood2024} will be Europe's first operational space weather satellite, positioned at the Sun–Earth L5 Lagrange point and offering an unprecedented side-view of the Sun enabling early detection of solar events, improve forecast lead times by up to four or five days, and support protection of critical infrastructure.\\ Other long-term missions such as the Advanced Composition Explorer (ACE), Geotail, IMP and Wind \cite{Stone1998, Wilson2021, Nishida1992} provide continuous measurements of solar wind and interplanetary magnetic fields near the L1 point. Networks of ground-based GNSS stations and radars offer extensive Total Electron Content (TEC) and plasma measurements \cite{Jakowski2011}. This diverse body of data spanning multiple platforms, temporal cadences, and modalities constitutes a heterogeneous observational corpus that is indispensable for both scientific discovery and operational applications.

With the growing availability of ionospheric observations, the field is becoming increasingly well-suited for machine learning (ML) applications. Yet a major bottleneck is the limited availability of machine learning ready datasets: existing products were not designed with ML workflows in mind. Data sources are heterogeneous in resolution and format, often sparse, and require extensive preprocessing before they can be used effectively for training and evaluation. This lack of standardized, ML-ready datasets slows progress and prevents systematic comparison of models.

To address this gap, we focus on building an ML-ready ionospheric dataset that integrates heterogeneous, multi-source observations, harmonizes temporal and spatial scales, and is tailored to the needs of data-driven modeling. This dataset provides the foundation upon which advanced ML architectures can be developed, tested, and benchmarked for global-scale ionospheric nowcasting and forecasting. Ultimately, creating robust ML-ready datasets is a necessary step toward building digital twins of the ionosphere and unlocking the full potential of data-driven space weather research.

\begin{figure}
    \noindent
    \makebox[\textwidth][c]{\includegraphics[width=1.4\textwidth]{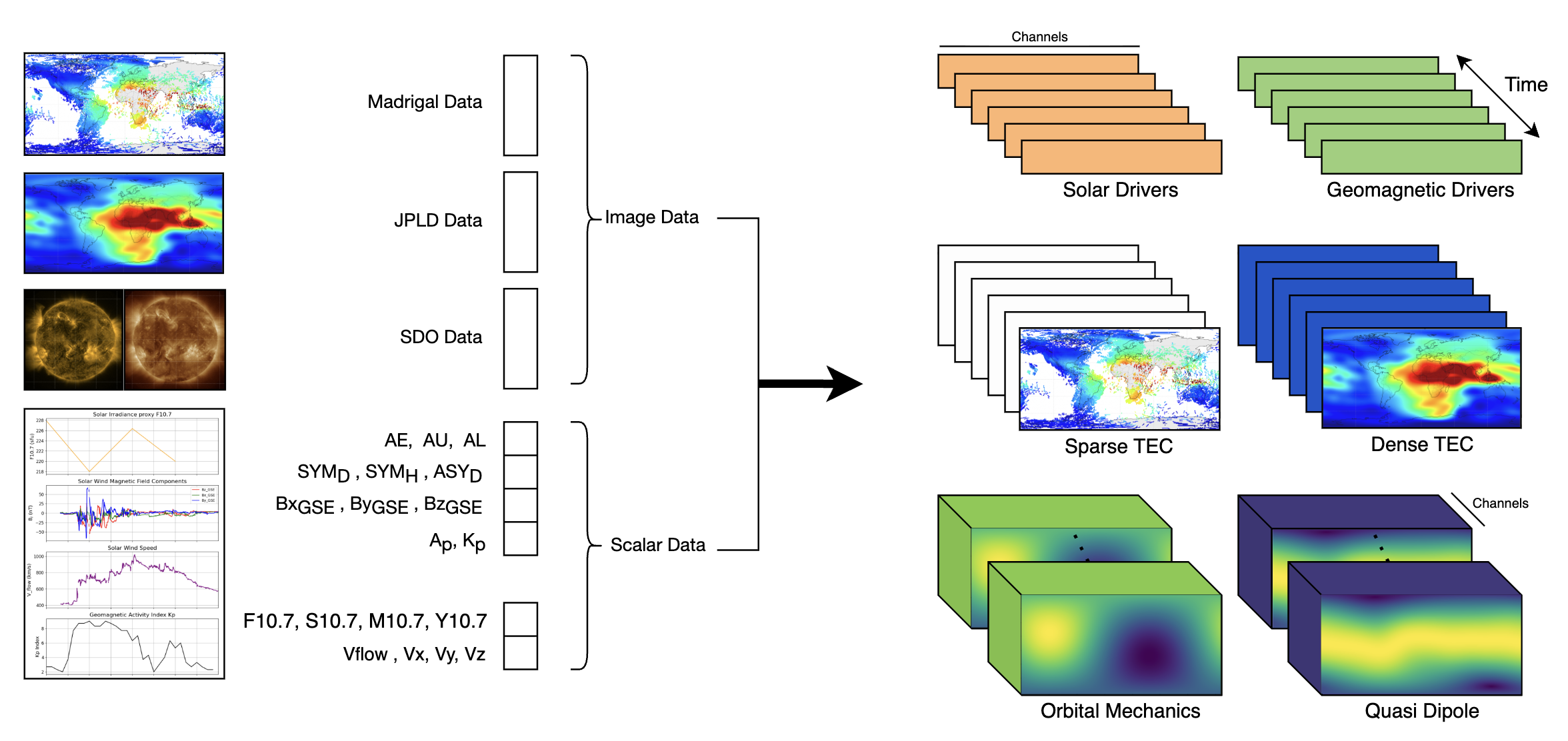}}
    \caption{Visualization of dataset inputs and alignment in time and dimension. Output dataset incorporates solar and geomagnetic driver data, sparse and dense TEC maps, and orbital mechanics and quasi-dipole data calculated over a latitude-longitude grid.}
    \label{fig:data-fig}
\end{figure}

\section{Dataset}

\begin{table}
\centering
\renewcommand{\arraystretch}{1.6}
\makebox[0pt][c]{
\begin{tabular}{p{3cm} p{5cm} p{1.5cm} p{2cm} p{4cm}}
\hline
\textbf{Source} & \textbf{Features} & \textbf{Cadence} & \textbf{Date Range}  & \textbf{Description}
\\
\hline
NASA/GSFC OMNI2 via OMNIWeb \cite{Stone1998, Nishida1992, Wilson2021} & 
AU, AL, AE (nT) \newline 
SYM-D, SYM-H, ASY-D (nT) \newline
$B_x GSE$, $B_y GSE$, $B_z GSE$ (nT) \newline
Solar wind speed, $v_x$, $v_y$, $v_z$ (km/s) &
1 min & 
2010-05-13 \newline - 2024-08-01 &
Geomagnetic indices related to magnetospheric/ionospheric currents and solar wind/IMF measurements. \\

NOAA SWPC/GFZ via Celestrak \cite{Matzka21}& 
Ap (nT) \newline 
Kp (no unit) &
3 hours & 
1997-01-01 \newline - 2025-10-12 & 
Geomagnetic indices describing the state of magnetospheric and ionospheric currents. \\

JPLD \cite{Mannucci1998, Martire2024, jpld} & 
Dense TEC GIM (1°×1° grid TECU) &
15 min & 
2010-05-13 \newline - 2024-07-31 &
TEC maps generated by JPL based on daily operational GIM processing for the Deep Space Network (DSN). \\

Madrigal \citep{Rideout} & 
Sparse TEC GIM (1°×1° grid TECU) &
5 min & 
2010-01-01 \newline - 2024-08-01 &
TEC maps derived from GNSS receivers, hosted by MIT Haystack Observatory. \\

SDO-FM \citep{walsh2024}& 
EUV irradiance embeddings (unitless) &
15 s & 
2010-05-13 \newline - 2024-08-01 &
Full solar disk irradiance observations, reduced with NVAE embeddings for ML applications. \\

Space Environment Technologies \cite{Tobiska2012} & 
F10.7, S10.7, M10.7, Y10.7 (sfu) \newline
JB08 dSt/dt (K) &
Daily & 
1997-01-01 \newline - 2025-10-12 &
Solar flux indices across multiple wavelengths; includes JB08 thermospheric heating rate. \\

Orbital Mechanics \cite{Giorgini1996} & 
Solar and lunar zenith angle (°) \newline
Subsolar and sublunar point (lat, lon) \newline
Solar and lunar antipode point (lat, lon) \newline
Earth–Sun (AU) and Earth-Moon (LD) dist. &
Variable & 
Variable &
Geometrical features derived from solar and lunar positions relative to Earth. Computed with the same temporal resolution as input features. \\

Quasi Dipole \cite{Laundal2017} & 
Latitude and longitude in quasidipole frame &
Yearly & 
2010-01-01 \newline - 2024-12-31 &
Yearly quasidipole maps: projection of Earth’s magnetic field lines into a geographic reference frame. \\
\hline
\end{tabular}%
}
\vspace{2ex}
\caption{Summary of data sources, their channels, cadence, date ranges available in the data product and descriptions. The date ranges for certain datasets were selected to match the date range of SDO \citep{walsh2024}.}
\label{tab:data_sources}
\end{table}

Our dataset aligns heterogeneous data from a diverse set of sources, encompassing multiple modalities, cadences, and start/end dates. Data sources include global ionospheric maps (GIMs) \cite{hernandez2009igs}, both with sparse and dense measurements, Solar Dynamics Observatory (SDO) extreme ultraviolet flux embeddings, solar driver data, and geomagnetic driver data. Our data product aligns these data sources in time and incorporates relevant orbital mechanics and quasi-dipole features. The aligned data product is visualized in Figure \ref{fig:data-fig}, and is publicly available on AWS S3 \footnote{{https://zenodo.org/records/18343833} and {https://github.com/FrontierDevelopmentLab/2025-HL-Ionosphere-dataset}}.

We align our dataset to the start and end dates of the SDO Foundation Model \citep{walsh2024}, which are 2010-05-13T00:00:00 to 2024-08-01T00:00:00. Our dataset will be available with multiple cadences according to the dataset sources detailed in Table \ref{tab:data_sources}. Multimodal data is processed in our codebase, provided publicly on GitHub\footnote{https://github.com/FrontierDevelopmentLab/2025-HL-Ionosphere}, which also handles the alignment in time and processing of data features. The dataset is structured and queried by time.\\ As an additional product, we provide an event catalog which uses a simple threshold on the Kp time series to identify periods of enhanced geomagnetic activity. In particular, this catalog divides the entire time interval into sub-intervals associated with a specific geomagnetic storm flag using the NOAA G-levels. This classification criteria also take into account the duration of the event period (see Table \ref{tab:mestici-scale}). A schematic view of the event distribution in the considered time interval is shown in Appendix \ref{fig:data-catalog}. This physics-based classification was prepared to ensure proper model validation and mitigate data leakage (where portions of the same geomagnetic storm event are scattered across training and validation sets) for the IonCast model \cite{ioncast}, which aim to accurate forecast TEC global maps for all geomagnetic conditions.

\begin{table}
\centering
\renewcommand{\arraystretch}{1.3}
\begin{tabular}{ c c c }
\hline
\textbf{Event ID} & \textbf{NOAA G-Level (Kp)} & \textbf{Duration (hours)} \\
\hline
G0H$\ell$ & $K_p < 5$ (Calm) & $\ell$ \\
G1H$\ell$ & 5 $\leq K_p <$ 6 (Minor)   & $\ell$ \\
G2H$\ell$ & 6 $\leq K_p <$ 7 (Moderate) & $\ell$ \\
G3H$\ell$ & 7 $\leq K_p <$ 8 (Strong)  & $\ell$ \\
G4H$\ell$ & 8 $\leq K_p <$ 9 (Severe)  & $\ell$ \\
G5H$\ell$ & $K_p \geq 9$ (Extreme) & $\ell$ \\
\hline
\end{tabular}
\vspace{2ex}
\caption{The geomagnetic storms catalog classification scheme. The event-ID combines NOAA G-levels (defined by Kp) with storm duration $\ell$ in hours. For example, G2H6 means an event that reached the G2 level lasting at least 6 hours.}
\label{tab:mestici-scale}
\end{table}

\section{Challenges}

A key challenge in constructing our dataset is the presence of missing values and inconsistent cadences across the underlying data streams, posing a challenge in temporally aligning the datasets. Different sources also adopt non-standard conventions for encoding missing values. For example, the OMNI dataset \cite{Stone1998, Nishida1992, King2005} marks gaps with sentinel values that differ over channels, making it difficult to detect. To standardize across datasets, we represent all missing values as NaNs. The OMNI dataset also contains certain features with multiple years of data missing. Any columns containing major gaps were removed. To deal with small holes, we use a simple forward-filling approach to fill in short gaps, using the most recent valid sample to fill NaN values. To determine whether to fill or skip a short gap, for each data stream, we define a maximum rewind time. For most data streams, the maximum rewind time is set equal to the native cadence of the dataset, to ensure only minor interruptions are filled. The one exception is OMNI, which has a rewind time of 50 minutes. If a gap exceeds the rewind time, the corresponding timestamps are skipped to avoid propagating stale data. The choice of rewind time is flexible and can be updated by the user if the default values are not suitable for the end user. This same forward-filling logic is also used as a simple interpolation strategy to resample all features to a standard temporal cadence.

\section{Baseline Results}

Our machine learning ready data product has enabled the training of a suite of global ionospheric forecast models showing promising results on autoregressive forecasts of TEC. These models are trained on a 15-minute cadence aligned data product, with dense TEC maps from JPL serving as prediction targets. Our models, named IonCast \cite{ioncast}, outperform baseline persistence TEC forecasts and produce accurate forecasts up to 12-hour lead times. The IonCast models include an LSTM \cite{Hochreiter1997} baseline model, a Spherical Neural Operator Model (SFNO) \citep{Bonev2025}, and a GraphCast \citep{Lam2023} model inspired by recent advancements in weather modeling. 

Our data product is available publicly on AWS S3, along with the codebase used to align, process, and split data based on geomagnetic storm events \footnote{{https://zenodo.org/records/18343833} and {https://github.com/FrontierDevelopmentLab/2025-HL-Ionosphere-dataset}}. Our codebase includes PyTorch datasets that prepare training data according to user-specified start and end ranges, dataset-specific normalization schemes, and example PyTorch model training code. 

The growing availability of data in the fields of space weather and heliophysics has made it possible to curate large datasets of heterogeneous data sources for machine learning model training. Yet, no existing resource aligns ionospheric TEC maps (both sparse and dense) with solar and geomagnetic driver data to enable ionospheric modeling with machine learning. To this end, we present a novel dataset that integrates data from diverse modalities, sources, and cadences into a single, machine-learning-ready product.

\section*{Acknowledgments}
This research is the result of the Frontier Development Lab, Heliolab a partnership between NASA, Trillium Technologies Inc. (USA), Google Cloud, NVIDIA and Pasteur Labs, Contract No. 80GSFC23CA040. A portion of research was carried out at the Jet Propulsion Laboratory, California Institute of Technology, under a contract with NASA. The authors thank Andrew Smith and Umaa D. Rebbapragada for their valuable insights, NASA's Goddard Space Flight Center, and NASA's Jet Propulsion Laboratory for their continuing support. 


\appendix


\pagebreak
\bibliographystyle{unsrt}
\bibliography{references}

\newpage
\section*{Appendix}

\begin{figure}[h]
    \noindent
    \makebox[\textwidth][c]{\includegraphics[width=1.2\textwidth]{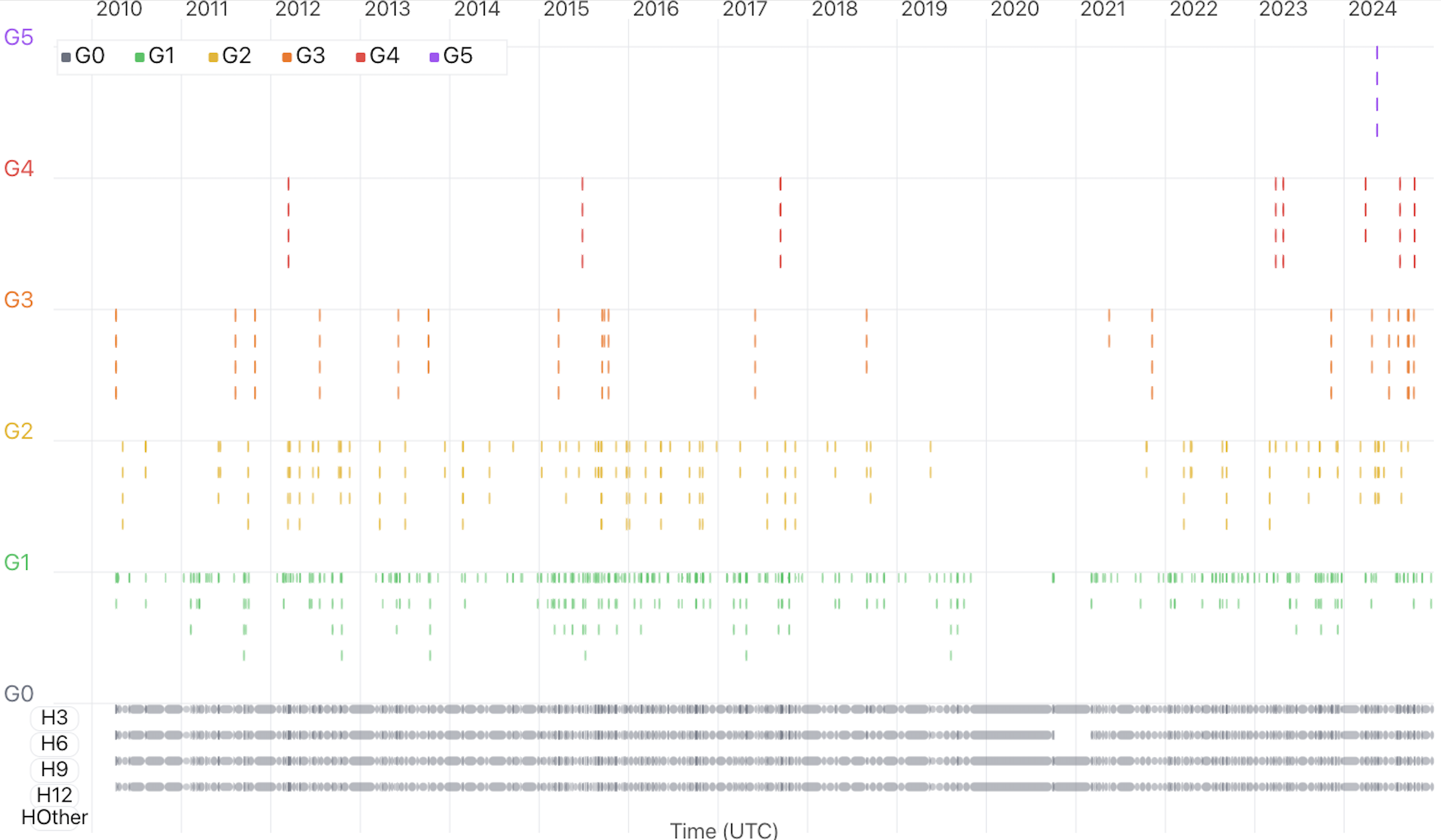}}
    \caption{Visualization of the 'Monitoring Event Space-weather TEC Ionospheric Catalog Index' (the MESTICI scale) showing temporal distribution of the Event class for the entire dataset time interval (2010-2024). The x and y axes represent the time (years) and the intensity of the event (G-level), respectively. Each class bin in the y-axis is then divided into four segments, which correspond to the event duration, as shown in the lower part of the plot.}
    \label{fig:data-catalog}
\end{figure}



\end{document}